\pdfoutput=1

\documentclass[11pt]{article}

\usepackage[]{acl}

\usepackage{times}
\usepackage{latexsym}
\usepackage{graphicx}
\usepackage{multirow}
\usepackage{xcolor}
\usepackage{enumerate}
\usepackage[shortlabels]{enumitem}
\usepackage{multirow}
\usepackage[T1]{fontenc}

\usepackage[utf8]{inputenc}

\usepackage{microtype}

\title{Human in the loop approaches in multi-modal conversational task guidance system development}

 \author{Ramesh Manuvinakurike, Sovan Biswas, Giuseppe Raffa, Richard Beckwith, \\
        {\bf Anthony Rhodes, Meng Shi, Gesem Gudino Mejia, Saurav Sahay},
        {\bf Lama Nachman}
        \\ first.last@intel.com
        \\ Intel Labs}

\begin{document}
\maketitle
\begin{abstract}
Development of task guidance systems for aiding humans in a situated task remains a challenging problem. The role of search (information retrieval) and conversational systems for task guidance has immense potential to help the task performers achieve various goals. However, there are several technical challenges that need to be addressed to deliver such conversational systems, where common supervised approaches fail to deliver the expected results in terms of overall performance, user experience and adaptation to realistic conditions.
In this preliminary work we first highlight some of the challenges involved during the development of such systems. We then provide an overview of existing datasets available and highlight their limitations. We finally develop a model-in-the-loop wizard-of-oz based data collection tool and perform a pilot experiment.
\end{abstract}

\section{Introduction}

In this work, we present the development of a conversational task guidance system. The goal of this work is to discuss the limitations of the current datasets and methods to fill the gap by developing a new dataset and method - based on human-in-the-loop approach - that helps address some of the issues. 
Such a system can find application in manufacturing, daily and do-it-yourself tasks (cooking, changing a flat tire, jump start a car, etc.), system maintenance, etc. Such systems can inform users about the next action to perform, 
tools to use, answer questions, provide insights about the current state, debug an issue, preemptively interject when an error occurs. 
Conversely, such a system can continually learn from the users to improve its performance and adapt to changing conditions (continual learning). Development of such systems is complex, including models for, but not limited to, knowledge retrieval, action recognition and segmentation, multi-modal dialogue understanding,  dialogue management, language generation, common-sense reasoning, co-reference resolution, and grounding. Data availability remains a challenge for the development of such systems, although numerous datasets exist to study these questions in isolation. 
While several datasets have been developed focusing on problems mentioned, a unified dataset for conversational task guidance is not yet available. Such a dataset should include, i) Visual demonstrations: for visual understanding, ii) Conversations relevant for task guidance: for training dialogue modules, iii) Ground-truth (document) for reference/retrieval, iv) Ontology of task structure: for reasoning and personalization and, v) Annotations for developing supervised models. 
We will now define the components of such a conversational task guidance system. 

\textbf{Task:} The conversations in a task guidance scenario is a dialogue between the task ``performer" and the ``guide''. The system acts as a task `guide' that helps the task performer realize the goal of a complex situated task. The performer acts as the ``human-in-the-loop'' of the system \cite{amershi2014power}, who helps the dialogue system components to be trained, fine-tuned and adapted based on the dialogue. 

\begin{figure}[!ht]
    \centering
    \includegraphics[width=\linewidth]{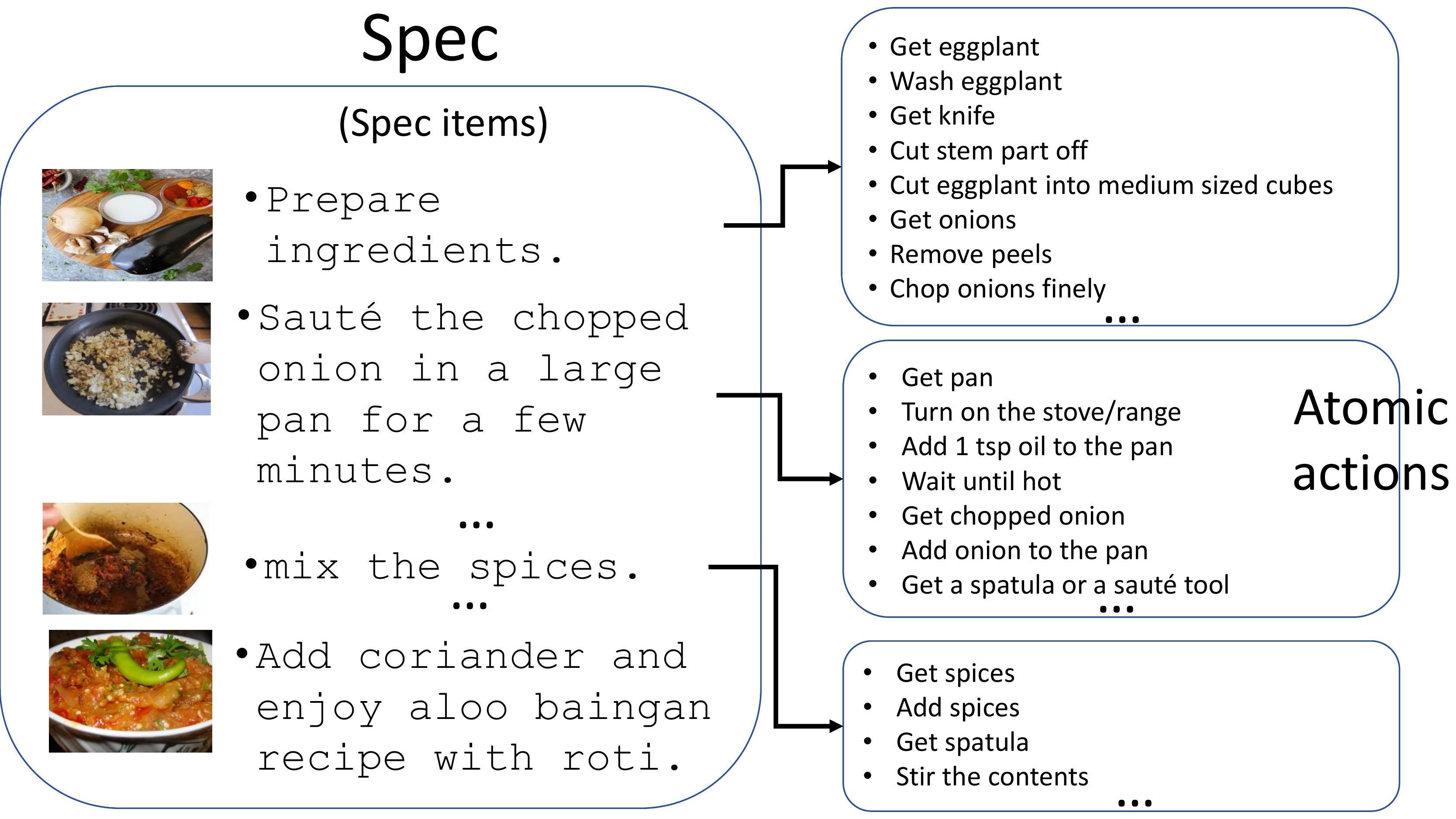}
    \caption{Shows the spec, spec items and the actions hierarchy for task guidance scenario. }
    \label{fig:spec_spec_items}
\end{figure}

\textbf{Spec, Spec Item \& actions:} To complete the task, the performer is expected to follow a ground-truth document called the `spec' (e.g., recipe) which provides details about the task. The spec consists ordered steps easily interpretable by human called `spec items'. Each step or a `spec item' typically consists of a set of `atomic actions' (referred as actions henceforth) that is needed to successfully complete the spec item. Note, these atomic actions are sometimes interchangeable or even optional (Figure~\ref{fig:spec_spec_items}). 
The guide (in this case the dialogue system) needs to recognize the actions performed by the performer in an online manner and simultaneously identify the spec item corresponding to the action.  
Thus the system is required to identify actions being performed by the performer at two levels of abstraction. 
This is an important aspect as the guide can then recommend the next spec item or action to the performer. 
The process of identification of actions corresponding to a spec item bears resemblance to commonsense reasoning \cite{forbes2019neural,losing2021extraction}. 

\textbf{Semantic frames:}
For representing a step in a process (spec item or actions), we use semantic frames which forms the understanding module. Table~\ref{tab:semanticframes} shows the slot and values in the semantic frame and the definitions. The semantic frames bear resemblance to the slot-value representation for dialogue state \cite{williams2014dialog}. However, in this task each spec item consists of multiple actions and are represented as semantic frames, each step (spec item) in the execution can be thought of a semantic frame consisting of multiple sub semantic frame (action). Such a hierarchy bears resemblance to the task of frame \cite{el2017frames} representation of a state tracking task. However, the distinguishing factor is the hierarchical representation of a process spec item as a series of optional action items whose sequence is probabilistic and that the semantic frames are grounded in visual demonstrations (videos). Such a representation of the task enables development of a multi-modal dialogue system for task guidance.  

The main contribution of this work is, i) Introducing a novel representation for a multimodal dialogue system development for task guidance, ii) data consisting of multiple sections for training the models discussed above, iii) A novel model-in-the-loop Wizard-of-oz framework for task guidance data collection. 


\begin{table}
$\left\{
\begin{tabular}{@{}l@{}}
    \textbf{Action}: Action term \\
    \textbf{Tool}: Object used for action       \\ 
    \textbf{Receiver}: Object receiving the action    \\ 
    \textbf{Location}: where action is performed    \\ 
    \textbf{Temporal}: temporal aspects of the action    \\ 
    \textbf{Direction}: direction of the action   \\ 
    \textbf{Manner}: how the action is performed \\ 
    \textbf{Extent}: until when the action is performed  \\ 
    \textbf{Purpose}: Purpose of the action     \\ 
    \textbf{Note}: Additional non-actionable notes        \\ 
\end{tabular}
\right\}$
\caption{Semantic frame used for understanding modules.}
\label{tab:semanticframes}
\end{table}


\begin{figure*}[t!]
\includegraphics[width=\linewidth]{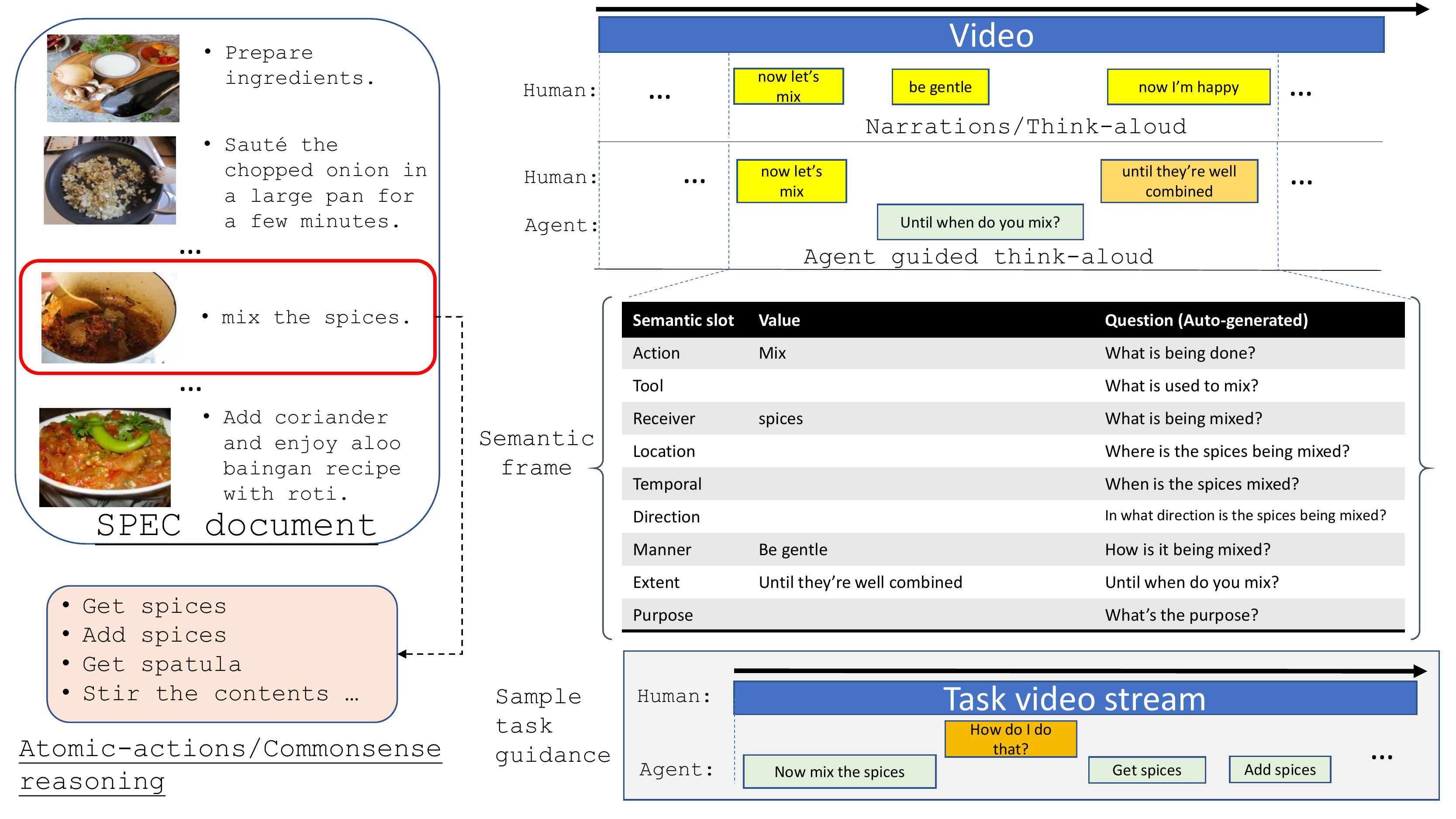}
\caption{Figure shows a sample data from our dataset. The spec document consists of instructional images and natural language instruction for carrying out the process to completion. Each process has accompanying videos with narrations.  } 
\label{fig:overallexample}
\end{figure*}

\section{Related work}

\textbf{Datasets:} 
Several datasets have been developed in recent years focused on the task of task guidance. We discuss a small subset of the relevant datasets. Activitynet \cite{caba2015activitynet} focused on developing a dataset to identify activities and the related hierarchies in general web videos. 
Youcook2 dataset \cite{zhou2018towards} consists of cooking videos from youtube consisting of annotated instructions. 
Annotation of the dataset is resource intensive. Various benchmarks have been developed on datasets with focus on unaligned videos and annotations (typically narrations or weak labels). 
Howto100m \cite{miech2019howto100m} dataset consists of more than 134k hours of videos and corresponding narrations collected from youtube intended to learn a joint embedding representation of video and text. 
COIN \cite{tang2019coin} dataset consists of >475 hours of instructional contents from youtube. 

Grounding the videos in a spec document has also garnered interest. Videos for video classification (MedvidCL) and question answering (MedvidQA) was introduced recently \cite{gupta2022dataset}. MedvidCL task involves classifying the video into medical instructional, non-instructional and others. MedvidQA involves answering questions by highlighting relevant section of the videos that answers the question. The authors in this work leverage wikihow (spec) during the process of data curation. Similarly, \newcite{colas2020tutorialvqa} have developed a question-answering dataset to highlight the video segment from tutorial videos that answer the question. 

One of the major challenges with the videos typically crawled from the web is that they're edited with the purpose of conveying the message as quickly, rather than to demonstrate the real-time execution of a particular task. This is important in the task guidance scenario, since the temporal aspects of the actions are a useful for the temporal aspects of actions which in-turn is useful for the planning problems. 
Dataset that capture the real-time egocentric have also been developed. EPIC-Kitchens \cite{Damen2018EPICKITCHENS,damen2022rescaling} consists of videos and narrations/annotations where users with a head-mountedgopro camera perform various tasks (e.g., cooking, cleaning) in a kitchen. EGO4D \cite{grauman2021ego4d} is a recent dataset consisting of users performing large number of daily activities. Recent assembly 101 \cite{sener2022assembly101} consists of videos performing assembling tasks with hierarchy of tasks also represented. 

\textbf{Semantic frames:}
Conversational task guidance systems require fine-grained multi-modal understanding abilities such that they can provide assistance relevant to the user. For understanding we use semantic frames representation \cite{traum2003semantics,devault2011incremental}. Our definition of semantic frames for task guidance scenarios is motivated by the semantic role labeling from propbank \cite{kingsbury2002treebank} annotation schemes. The semantic roles of a given sentence are verb centric, where each verb in a given sentence is labeled with arguments such as patient, agent, benefactive etc. The task of identifying such semantic roles from the visual context (image or video) has been studied \cite{gupta2015visual,yang2016grounded,sadhu2021visual}. In our work, we define a semantic frame where the arguments are centered around actions of interest, rather than verb/auxiliary as present in the propbank annotation scheme \cite{bonial2012english}.

\textbf{Retrieval:}
The process of retrieval in a task guidance domain, specifically cooking domain has been studied as a problem involving retrieving images from recipes/ingredients or as a problem of generating recipes from the images \cite{salvador2017learning,salvador2019inverse}. 
In our work, one of the focus is to highlight the spec items incrementally as the demonstrations occur in real-time. The spec is authored by experts, and this can be considered a ground truth document that needs to be aligned to the demonstrations.  
Closest to our work is that of Multimodal-aligned-recipes \cite{lin2020recipe} where the authros construct a dataset of youtube videos where the segments from the videos are automatically aligned to recipes and recipe items. This is however not incremental, which is a requisite for our work. For building the models for spec retrieval, we explore recent contrastive learning models which have given promising results \cite{chen2020simple,he2020momentum,radford2021learning,xu2021videoclip,jia2021scaling}. 


\textbf{Conversational task guidance systems:}
Task guidance system development has been of interest for sometime. Such systems have been developed to help navigate through operation of daily household equipment \cite{rich2007diamondhelp}, medical domains helping conduct surgery \cite{escobar2020review,tillou2016robotic}, cooking for differently-abled \cite{ramil2021allergic} etc. Given the complex nature of the recent works have focused on a tractable set of problems such as question-answering \cite{anantha2020open,ramil2021allergic,feng2021multidoc2dial}. It is important to note the distinction between task guidance systems discussed in this work with systems leveraging human guidance \cite{zhang2021recent} for learning. We believe such methods to learn from human demonstrations as an important characteristic of the task guidance systems. 

\textbf{Data collection:}
The development of such systems is challenging and requires large amounts of annotated data. Following \newcite{Damen2018EPICKITCHENS,grauman2021ego4d} we follow the notion of developing the data with weakly aligned human narrations as labels for visual data. For building understanding models for semantic frames extraction, given the promising aspect of controlled crowd-sourcing, we're exploring the development of a corpus leveraging crowd-sourcing paradigm \cite{he2015question,roit2020controlled}. we believe Question-answering SRL paradigm of leveraging crowd-workers to answer the questions given a context to fill semantic frames as a promising direction \cite{roit2020controlled}. 

\section{Data}

The task guidance we are interested in this work involves humans performing a task where they're expected to follow a \textbf{spec} document. It is important to note that the spec document is authored by a task expert for the human task performer and such documents are typically available during task performance. Such document assumes prerequisites (e.g., in recipes to understand 'mince garlic' the human performer needs to know how to do this) that might not all be interpretable by novice humans or machines. Hence, we develop a task breakdown structure where items in the spec document is broken down into atomic actions which a novice can interpret (it is important to note that it is possible that a task performer cannot still understand these). These atomic actions can also be represented as semantic frames. 

\textbf{Think-aloud/narrations} have gained popularity (for example \newcite{Damen2018EPICKITCHENS,miech2019howto100m,tang2019coin}) since they can be leveraged as noisy weak labels overcoming the annotation phase which can be expensive. Such noisy labels have yielded promising results in various vision based tasks \cite{radford2021learning,jia2021scaling,xu2021videoclip,wang2021actionclip} outperforming models trained explicitly on limited annotated samples. 
The main advantage of such an approach is that the narrations can be provided by novice annotators who don't require additional training. However, noisy labels due to inconsistent vocabulary usage and misalignment between the visual frames and narrations can be a drawback. These drawbacks can be overcome potentially by carefully designing self-supervision pretext tasks to learn a joint representation  \cite{xu2021videoclip,wang2021actionclip}. 
One of the ideas we are currently exploring that has yielding promising preliminary results is the `conversational system guided think-aloud/narrations'. We collect these conversational system guided narrations data collection either `post-hoc' or `during' task demonstration. We leverage wizard-of-oz approach to collect the narrations data for think-aloud/narrations data post-hoc and during task demonstrations.

In the \textbf{`post-hoc'} system, we assume that there exists large amounts of videos without any narrations. 
To collect the weak labels, we make the users (typically task experts) watch the pre-recorded videos post-hoc and narrate them. When the users are not sure what to narrate or narrate incompletely (for e.g., don't narrate the action or objects) the conversational system assists them by generating questions specific to slots in the semantic frames. 
This is challenging since additional time associated with question asking could further exacerbate the problem of alignment. To help mitigate this issue, we need capability for the dialogue system to have video manipulation abilities (play, pause, rewind, forward, loop, zoom) and collect narrations from the users specific to the video frames. Although this approach doesn't solve the problem of alignment, it could potentially mitigate some of the issues.
We are also interested in developing the dialogue system for collecting the narrations \textbf{`during'} the task performance. This phase is important as it can accelerate the system development by decreasing the time involved in collecting the narrations post-hoc. However, there are additional challenges involved in deploying such systems. The models need to operate online and incrementally. The goal for the system to help collect the weak labels during task demonstration is to, i) prod the performers to provide narrations, ii) if information is missing in the narrations (slots in semantic frames) ask questions to fill the incomplete slots, iii)  highlight the relevant spec item for the task demonstrator to collect feedback about the errors in retrieval model predictions. 

\section{Wizard system}

\begin{figure*}
    \centering
    \includegraphics[width=\linewidth]{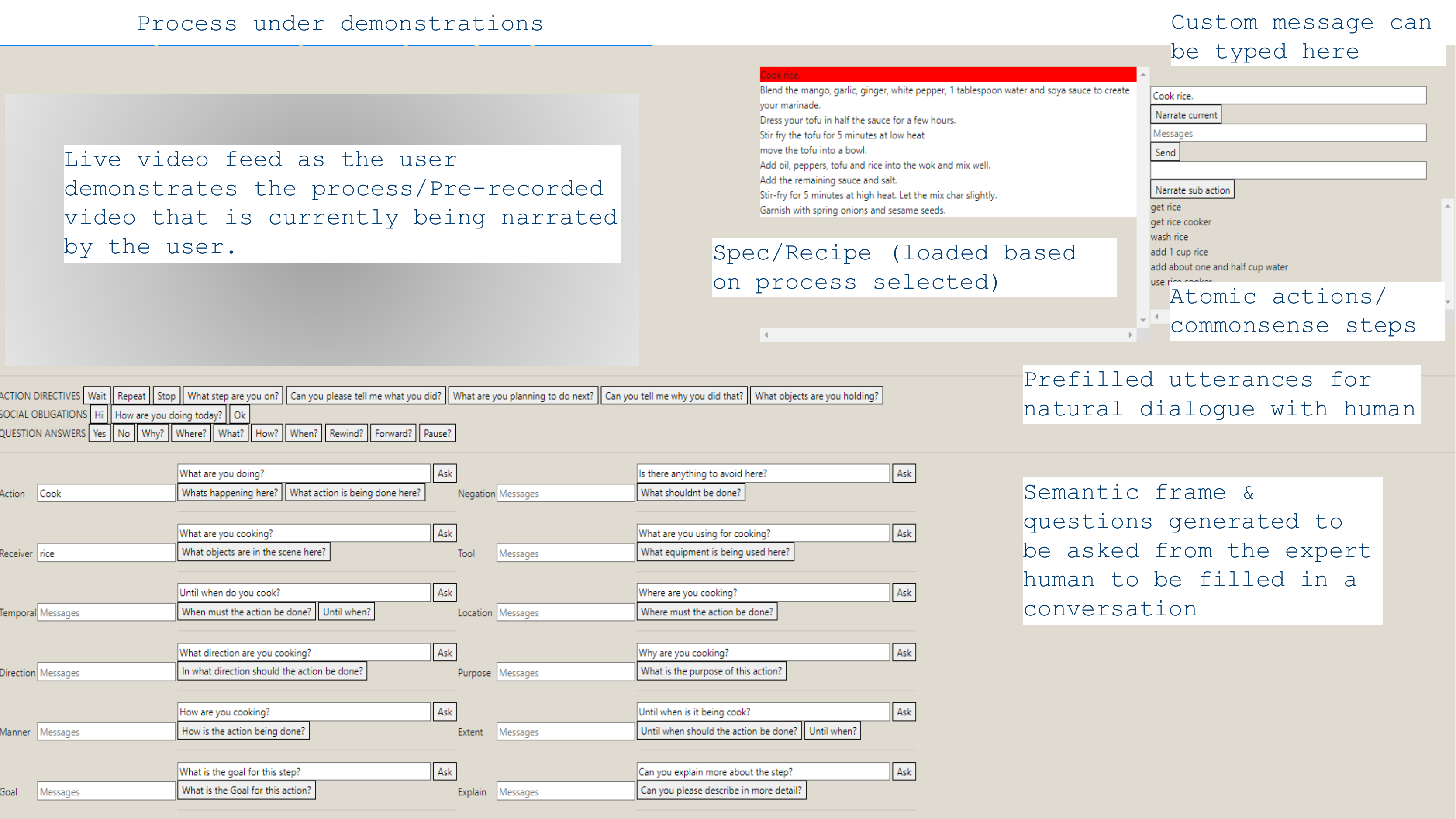}
    \caption{Shows the wizard interface for collecting the think-aloud/narrations. The wizard system backend is designed to interact with the semantic frame extraction models, question generation model, spec-retrieval and action recognition models. }
    \label{fig:wiz_interface}
\end{figure*}

The wizarding system was developed to collect spoken conversational data between the human narrator (post-hoc)/task performer (during) and the agent. The wizard interface and the user interface for both the `post-hoc' and `during' task data collection is similar in design (Figure~\ref{fig:wiz_interface}). The only difference being that the video section is either recorded or live streamed. The wizard interface contains pre-filled utterances that can be played back to the user and is designed to aid the goal of collecting more relevant narrations. 
The wizard also has access to the predictions from semantic frame extractor that can assist in asking relevant questions to aid better narrations from the user. The semantic frame extractor is a BERT based \cite{devlin2018bert} model. 
The questions are also pre-generated using BART based sequence to sequence model \cite{lewis2020bart}. 
The wizard can either choose to use these questions as is, edit them or choose to enter a new question. The atomic-actions and an image representing each step are provided manually for every item in the spec document by the expert.

\begin{figure}[!ht]
    \centering
    \includegraphics[width=\linewidth]{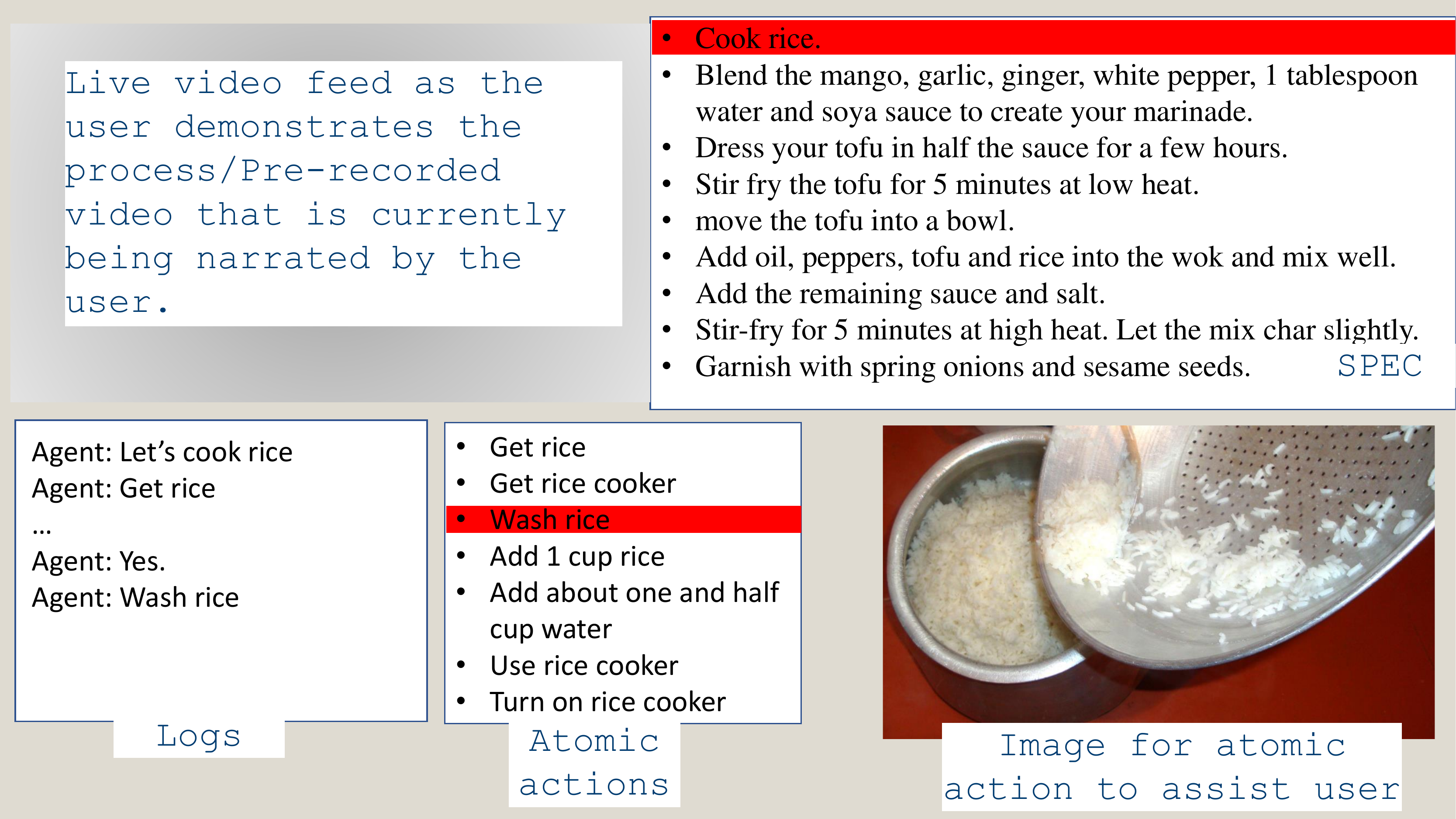}
    \caption{Shows the user interface for collecting the think-aloud/narrations. The user system shows the live video feed, spec, spec item current under execution, atomic action current under execution, conversation logs and the image corresponding to the atomic action to aid the task performer in execution of the task. }
    \label{fig:user_interface}
\end{figure}

The wizard in the `post-hoc' phase controls the video playback and asks the narrator to describe the video by selecting relevant buttons on their screen. The wizard's selection is played back to the user using TTS (Text to speech). The user's narration is relayed to the wizard screen live during the interaction using webRTC. The conversations are recorded on the server with timestamps. We'll now briefly discuss the models developed for semantic frame extraction, action recognition, spec item retrieval and question generation. 


\section{Models}

The models have access to video frames from the beginning of the execution until current n-th frame. $F = \{F_0, F_1, \ldots, F_n\}$. The spoken narrations by the users are available as transcription chunks $T = \{T_0, T_1, \ldots, T_p\}$. These transcription chunks are extracted from the ASR (Automatic speech recognizer). Each task consists of m spec items $S = \{S_0, S_1, \ldots, S_m\}$. Each spec item correspond to a system state. It is ensured that each spec item corresponds to a single semantic frame. These semantic frames may be incomplete. 

\subsection{Action Segmentation and Recognition:}
The objective of the action segmentation and recognition model is to recognize actions being performed at any frame $F_t$ at the $t^{th}$ time in the online manner. Each action has long temporal temporal dependency. For example, an action `wash rice' is temporally dependent on `get rice' action as shown in Fig. \ref{fig:user_interface}.
Thus, we use MSTCN \cite{farha2019ms,li2020ms} that recognize and segment actions based on sequence of frame features. A frame feature $F_t$ at time $t$, in our model, is captured using 3DCNN such as Slowfast \cite{feichtenhofer2019slowfast} over a sliding window of 64 frames. A standard MSTCN requires the entire video to perform action segmentation and recognition, which is a key restriction in our online case. Thus, we modify the individual temporal convolutions of MSTCN to causal temporal convolutions \cite{van2016wavenet} that only intakes all the frame features up to time $t$ to predict the action. Due to the use of the causal convolutions, the model is suitably modified for the online nature of the action recognition.

\subsection{Spec item matching}

The problem of spec item matching involves utilizing the current video frames, $F$ and/or the narrations, $T$ (if available) and estimate the current spec item $S_i$ that is `currently' being performed at a cadence as selected by the user. This is important to the task performer as they can use this information to plan the next set of actions. For baseline we use the models in a zero shot manner, i.e without fine-tuning the models. 

\textbf{Using transcriptions:} We use sentence transformers as text encoders to get the text embeddings from the transcriptions $T$ and spec items $S$. We use sentence BERT\footnote{sentence-transformers/all-MiniLM-L6-v2} \cite{reimers-2019-sentence-bert} to get the sentence embedding. The semantic similarity between the narrations and the spec items are calculated using cosine similarity metric. 

\textbf{Using video:} We use CLIP\footnote{https://huggingface.co/openai/clip-vit-base-patch16} \cite{radford2021learning} for the baseline model. CLIP model during inference takes as input an image ($F_j$) and text candidates ($S$) and ranks the candidates based on cosine similarity between the text embeddings and vision embeddings generated by the text and vision encoders of the CLIP model. 
We input each frame $F_j$ and get the similarity scores for each $S_i$. The $S_i$ with the highest aggregate similarity score for last `n' frames is selected as the best matching candidate. We find that this approach yields promising first results. There are several limitations with this approach, i) We find that image encoders from CLIP yield better results than using video encoders. However, use of video encoders self-supervised training could potentially yield better results. We leave this to the next steps.  ii) Since, we aggregate the results (irrespective of image or video encoders) there will be a lag in displaying the current spec item. A segmentation model could potentially help resolve this issue. We leave this to the future work and point to the importance of such segmentation models. 

\subsection{Semantic frame extraction}

The semantic frame extractor (SFE) module is a token classifier module which extracts the semantic frame from the input text ($S$ or $T$). The SFE module is adapted from the SRL (Semantic Role Labeling) BERT model \cite{shi2019simple} trained on Ontonotes dataset \cite{pradhan2013towards}. We use this SRL BERT and stack a linear layer mapping the original labels into the semantic tags definitions shown in Table~\ref{tab:semanticframes} in this work. An expert provides a map from the original propbank labels into the  new semantic frames (Example: $B-Verb \rightarrow B-Action$, $I-ARG1 \rightarrow I-Tool$ etc.). These mapping can then be used to initialize the weights of the newly stacked linear layer with 1 if the mapping is true, 0 otherwise. One of the earlier works applying this approach of rule integration was performed by mapping the input rules using the bag of words model \cite{hatzilygeroudis2004integrating}. We apply similar approach at the newly stacked linear layers. While this new layer is initialized with the human provided mapping, they still can be further finetuned with new data. 

Currently, the SFE is stateless, meaning it is not cognizant of the previously extracted incomplete frame which might be a useful to distinguish the semantic entities(e.g., Tool vs Receiver when the user has uttered only a referent). We believe this is another interesting area for future work. 

\subsection{Question generation}
The question generator is a trained sequence to sequence model that is trained using the BART model architecture \cite{lewis2020bart}. The model converts the input semantic frame entities and a prompt (slot for which question needs to be generated) into a question. We leverage QA-SRL dataset for \cite{klein2020qanom} for training our initial question generation model for filling a semantic slot. We preprocess the data to fit our needs in this work. 

\begin{figure}[!ht]
    \centering
    \includegraphics[width=\linewidth]{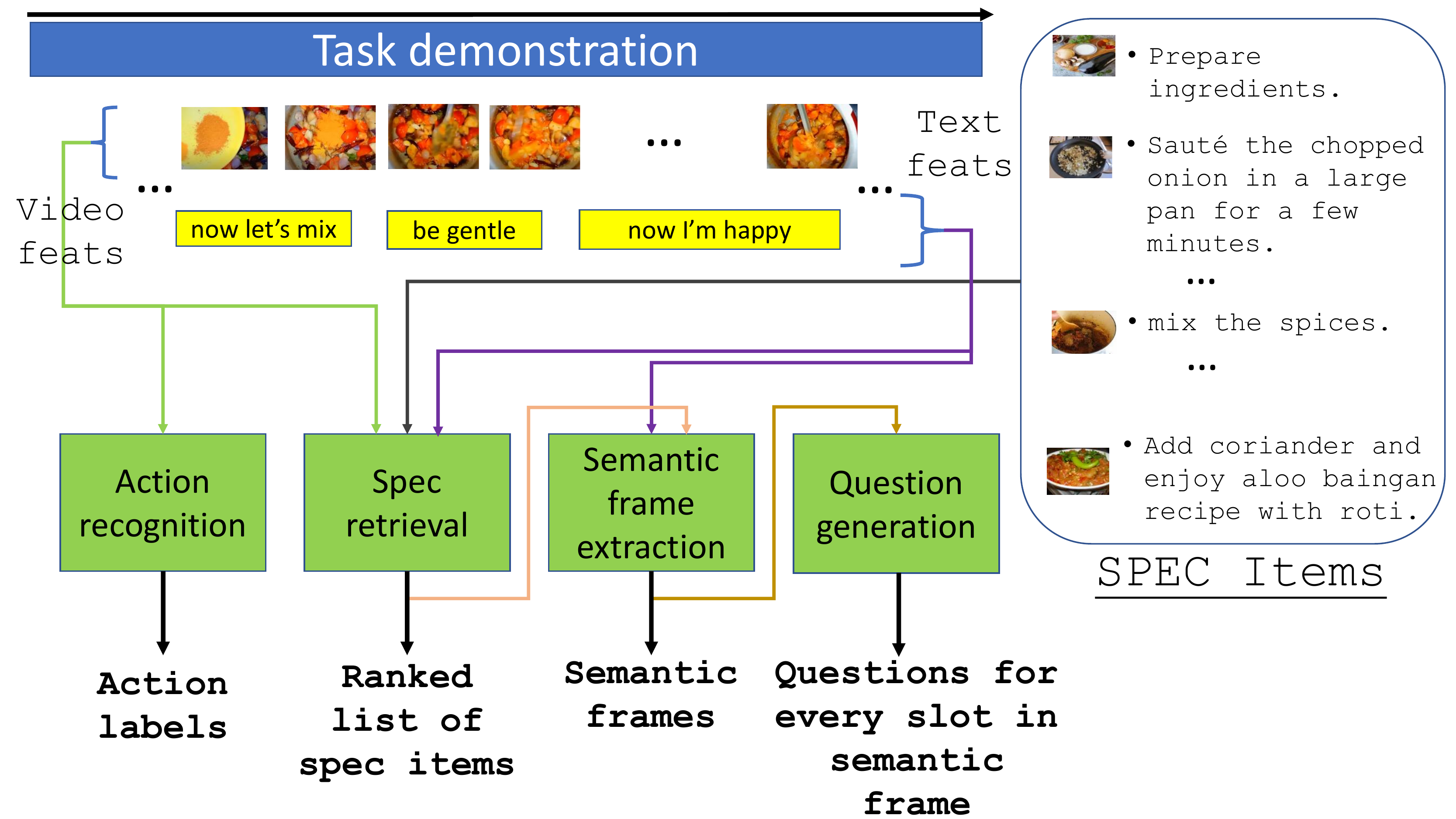}
    \caption{Shows the models and the inputs to the models developed in this work. }
    \label{fig:model_overview}
\end{figure} 

\section{Preliminary Results, Discussions \& Future work}

The wizard's role is not straightforward and if untrained, is cognitively burdensome. 
The wizard needs to perform various tasks in both the system guided think-aloud and task guidance. 
During the system guided think-aloud phase (post-hoc \& during) the wizard's goal is to elicit as much narrations from the user as possible, balancing the usefulness of the information retrieved and the user experiece of the task performer. 
This is a subjective goal designed to collect more text labels for a video and not helpful if the text doesn't help improve the existing models in the loop. 
Hence, the wizard is provided with the output from SFE models-in-the-loop which helps them choose specific question to ask to fill a particular semantic slot. 
The context visibility (task under execution (spec \& spec item), semantic frames, atomic actions) to the wizard is an important consideration as the goal is to collect collect relevant text utterance. 
Since we started the data collection without an in-domain data for training we report initial results on related datasets. 
This gives the baseline performance of the models and helps us compare the results as we collect data using our wizard interface. 

\textbf{Spec retrieval}

Table~\ref{tab:spec_item_retrieval} shows the model performance of spec item retrieval using the CLIP and Sentence BERT models every 1 second of the video. When utilizing the frame features $F$, we experiment with including a window size of `n', where `n' from the from the past is used to visual features extraction. 
We observe that the CLIP model outperforms the sentence BERT model. This can be attributed to the misalignment between the narrations and the video demonstrations. The spec item was annotated in the video using the visual demonstrations and not the narrations (e.g., person in the video demonstrations says `I'll be using jack' with the video focused on the person's face. The person then proceeds to get the jack without narrating anything. The annotations are corresponding the visual demonstrations and not the narrations).
The narrations from the users are spoken (semantic similarity model is trained on written text) and thus contains various disfluencies.
We also see that increasing the window size (i.e increasing amount of visual context) helps until certain point (n=6s). We further investigated individual tasks (performing CPR, changing tire, jump starting a car, making coffee using moka pot, repotting a plant) from Instructional videos dataset  \cite{Alayrac16unsupervised} and found that the performance varies across tasks. This is primarily because each task is unique and the duration of each step (corresponding to a spec item) varies depending on the nature of the task (e.g., chopping onion corresponds to different lengths depending the amount of onions being used in a recipe). It is thus important to include process information when modeling the spec retrieval for a given task. As a next step we plan to learn a joint vision and text encoder. 
When the narrations/transcriptions are absent it is not possible to run the time window analysis using the text. 

\begin{table}[]
    \centering
    \resizebox{\columnwidth}{!}{
    \begin{tabular}{l|l|l|l|l|}
         & R1 & R2 & RL & Acc \\ \hline
         SBERT & 39.37 & 16.324 & 26.89 & 19.14 \\ \hline
         CLIP (n = 1s)	& 43.51 & 26.28 & 33.14 & 31.6\\ \hline
         CLIP (n = 2s)	& 53.94 & 34.94 & 42.26 & 41.68\\ \hline
         CLIP (n = 4s)	& 53.84 & 34.79 & 42.56 & 41.91 \\ \hline
         CLIP (n = 6s)	& 55.20 & 36.51 & 44.16 & 44.35\\ \hline
         CLIP (n = 8s)	& 53.05 & 33.38 & 41.45 & 42.25 \\ \hline
    \end{tabular}}
    \caption{Shows the spec retrieval performance (ROUGE scores and Accuracy) on videos 50 videos Instructional videos dataset \cite{Alayrac16unsupervised}. The videos were annotated by spec documents for the videos obtained from wikihow. The videos belonged to 5 types of tasks (performing CPR, changing tire, jump starting a car, making coffee using moka pot, repotting a plant). } 
    \label{tab:spec_item_retrieval}
\end{table}

\textbf{Semantic frame extraction}
We find that on a small test set created from instruction videos spec data (50 sentences) we find the average precision=0.39 and recall=0.29. It is important to note that, although the tags for the semantic frames were motivated from the propbank labeling scheme there are numerous differences (for instance, SRL frames are created for every predicate (verb \& AUX) while the semantic frames in our corresponds to an action performed).

We finally collected 10 interactions (5 for post-hoc think-aloud/narrations, 2 during think-aloud narrations, 3 task guidance) interaction between the wizard and the user for the task of `cooking'. The total audio collected spans 3 hours. The users in all the scenarios successfully completed the task. We're currently in the process of collecting and analyzing the dataset. 

\textbf{Task guidance dialogue system}
A task guidance dialogue system interacts with the task performer and assists them in performing a given task.
A task guidance dialogue system is comprised of followed features,
i) Multi-modality: The system monitors the actions and objects present in the environment. This contains action recognition and object recognition tasks.
ii) Mixed initiative: The system can interject and provide corrections to the task performer if it predicts that the task is not being carried out satisfactorily. The system can also ask questions if it is uncertain about the task execution. 
iii) Document grounded question-answering: The system needs to be able to answer questions about the objects, actions, point the users to the next steps to be performed etc. The answers provided by the agent must be grounded in the documents/specs which provide `ideal' execution protocol of a given process. 
iv) Execution flexibility: While, the spec document provides the ideal execution of a given process, it is by no means the only way in which a task can be completed successfully. For instance, in a cooking recipes dataset if the spec document mentions chop the onions and then boil pasta, there is no impact on the outcome if the task performer chooses to put pasta for boiling before chopping the onion. However, the order of ingredients used for sauteing could impact the outcome. The system needs to account for the flexibility in task execution. 

\bibliography{anthology}
\bibliographystyle{acl_natbib}

\appendix



\end{document}